\documentclass{article}

    \PassOptionsToPackage{numbers, compress}{natbib}

\usepackage{ml4eng_2020}

\usepackage[utf8]{inputenc} 
\usepackage[T1]{fontenc}    
\usepackage{hyperref}       
\usepackage{url}            
\usepackage{booktabs}       
\usepackage{amsfonts}       
\usepackage{nicefrac}       
\usepackage{microtype}      
\usepackage{lipsum}
\usepackage{amsmath,amssymb,amsfonts, epsf}
\usepackage{breqn}
\usepackage{algorithm}
\usepackage{algorithmic}
\usepackage{mathtools}
\usepackage{graphicx}
\usepackage{cases}
\usepackage{color}
\usepackage{float}
\usepackage{amssymb}
\usepackage{appendix}
\graphicspath{{Fig/}}
\usepackage{nomencl}
\usepackage{mathtools}
\usepackage{booktabs}
\usepackage{nicefrac} 
\usepackage{microtype}
\usepackage{subcaption}
\usepackage[numbers]{natbib}

\usepackage{cleveref}

\title{Battery Model Calibration with Deep Reinforcement Learning}

\author{%
Ajaykumar Unagar \qquad Yuan Tian \qquad Manuel Arias-Chao \qquad Olga Fink\\[0.1cm]
  ETH Zurich \\
  \tt\small aunagar@ethz.ch \qquad \{tian,arias,fink\}@ibi.baug.ethz.ch
}

\begin{document}

\maketitle

\begin{abstract}
Lithium-Ion (Li-I) batteries have recently become pervasive and are used in many physical assets. 
To enable a good prediction of the end of discharge of batteries, detailed electrochemical Li-I battery models have been developed. Their parameters are typically calibrated before they are taken into operation and are typically not re-calibrated during operation. However, since battery performance is affected by aging, the reality gap between the computational battery models and the real physical systems leads to inaccurate predictions. 
A supervised machine learning algorithm would require an extensive representative training dataset mapping the observation to the ground truth calibration parameters. This may be infeasible for many practical applications. In this paper, we implement a Reinforcement Learning-based framework for reliably and efficiently inferring calibration parameters of battery models. The framework enables real-time inference of the computational model parameters in order to compensate the reality-gap from the observations. Most importantly, the proposed methodology \textbf{does not need any labeled data samples}, (samples of observations and the ground truth calibration parameters). Furthermore, the framework does not require any information on the underlying physical model.
The experimental results demonstrate that the proposed methodology is capable of inferring the model parameters with high accuracy and high robustness. While the achieved results are comparable to those obtained with supervised machine learning, they do not rely on the ground truth information during training.

\end{abstract}

\section{Introduction}

Recent advancements in Li-I batteries have increased their usage in various applications ranging from electric vehicles to drones and space exploration. Particularly for autonomous systems, it is essential to plan the missions reliably which requires an accurate prediction of the End-of-Discharge (EOD) time for the batteries. However, the currently available battery models \cite{comparebatteryHinz2019} suffer from an increasing uncertainty in their EOD predictions over time. This is mainly due to the degradation processes in the battery. The relationship between battery age and the discharge time is non-linear. Hence, sophisticated modeling techniques are required to estimate Battery Degradation Parameters that would be required for the EOD time prediction.

Previous work on Battery aging \cite{ning2004cycle, ning2006generalized, peterson2010lithium, barre2013review} have focused on the understanding of the aging process. Some data-driven methods based on empirical, probabilistic, and learning-based models \cite{saha2009modeling, daigle2016end, nuhic2018battery, andre2013comparative} have been proposed for Battery End-of-Life (EOL) prediction or State of Health (SOH) determination. However, these methods suffer from a strong dependence on labeled data, or are computationally expensive. These shortcomings limit their applicability in many real-world problems.

Battery aging models enable to track the aging process over time. The model parameters are then inferred from the empirical observations. This is also referred to as model calibration. Previous works have focused on traditional approaches like Kalman Filter \cite{andre2013comparative, bole2014adaptation}, Particle Filter \cite{saha2009modeling}, or explicit degradation models (Model-Based Prognostics) \cite{daigle2016end}. Kalman Filter and Particle Filter approaches do not require an underlying model, however suffer from a high computational burden during application time. Model-Based approaches assume the underlying model of the aging process. In this work, we solve the Battery Model Calibration process using Reinforcement Learning (RL) method, which can work in real-time and does not require an underlying model. 

Recent developments in model-free Reinforcement Learning have been applied to various control problems in Robotics \cite{bucsoniu2018reinforcement, kumar2016learning, deisenroth2011learning,han2019h_}, Water Systems management \cite{bhattacharya2003neural}, Computational Biology \cite{treloar2020deep}, and AutoML\cite{tian2020off}. Model-free RL methods have multiple advantages over traditional methods: (1) RL framework is highly general, the agent can learn to solve tasks without any knowledge of the underlying model. In our case, the agent can learn to calibrate the underlying parameters without any labeled data. (2) The policy learned via RL is robust to model uncertainty. (3) RL methods provide almost real-time performance since it only requires to evaluate the learned policy. Hence, Reinforcement Learning is a compelling alternative to other data-driven methods for Battery Model Calibration.

In this work, we propose to define the Battery Calibration problem as a tracking problem defined by a Markov Decision Process (MDP) and solve it with the Lyapunov-based Maximum-Entropy Reinforcement Learning algorithm. Specifically, we use Lyapunov-based Actor-Critic (LAC) method \cite{han2020actor,tian2019model} to provide stable tracking of the parameters. We use the Battery Model from the NASA prognostic model library \cite{daigle2013electrochemistry, nasaBattery} to simulate our RL environment. It is important to clarify that this library models the physical process of the discharge but not the battery aging process, which is our main focus here. To the best of our knowledge, ours is the first method applying Reinforcement Learning for the battery model calibration \footnote{In this work we use ``Model Calibration'' and ``Degradation Parameter Estimation'' terms interchangeably}.

\section{Related Work}

Model Calibration is essential in many engineering applications relying on simulations. As most of the systems undergo changes over time, physical model parameters need to be re-calibrated. If possible, this calibration process can be done offline by applying a reference loading condition and correlating the observed output with the expected one. However, in situations where it is not possible to calibrate the systems offline, online calibration is crucial.
    
For operational purposes of Batteries, it is important to have an accurate estimate of the EOD time. There are accurate discharge models for Li-I batteries \cite{daigle2013electrochemistry}. These models work based on the physical principles of the discharge process. However, such models are not able to estimate the degradation parameters of the Battery that change over time, and hence battery model real-time calibration becomes essential. 

There are three primary ways to assess Battery degradation parameters. (1) Direct estimation from observations (2) Bayesian Tracking principle (3) Model-Based Prognostics based on an explicit aging model. 

Firstly, in direct estimation methods, observations are used to learn the mapping from the battery outputs to the degradation parameters. For, example authors in \cite{nuhic2018battery}  used Support Vector Machine (SVM) model to learn this mapping. There are approaches that use Structured Neural Networks (SNN) \cite{andre2013comparative} to exploit knowledge of the degradation process. Such approaches show promising results in certain scenarios where we can obtain paired samples of observations and degradation parameters. 

Secondly, Methods based on Unscented Kalman Filter (UKF) \cite{bole2014adaptation} tracks the internal battery state to reduce the observation gap between predicted and actual output. Similar to this, Extended Kalman Filter (EKF) \cite{andre2013comparative} also tracks the internal state of the battery but with a different model for the discharge process. Such tracking algorithms provide model-agnostic parameter estimation. However, these methods are computationally expensive at the application time and suffer from a drift in parameter tracking. 

Thirdly, model-based prognostic methods assume an underlying degradation model for the aging parameters as the function of its usage. Authors in \cite{daigle2016end} used system identification techniques to estimate the parameters of the degradation model. These techniques provide accurate estimates as long as the physical degradation process follows the assumed model. Reinforcement Learning provides an alternative solution to these approaches while relaxing some of the constraints. Especially, in the scenarios where labeled data is not available, RL can learn from the observations and infer the model parameters.

With deep function approximators and sophisticated exploration techniques, Reinforcement Learning methods have made some significant progress in recent times. In our work, we focus on model-free RL methods based on Actor-Critic (AC) approach \cite{sutton2018reinforcement, konda2000actor}. Actor-Critic methods provide a framework for generalized policy iteration algorithms in which two networks (Actor and Critic) are updated continuously. Especially, Maximum Entropy-based RL formulation such as Soft Actor-Critic (SAC) \cite{haarnoja2018soft, haarnoja2018soft2} algorithms have shown good performance in different applications \cite{wu2020battery, li2020video}. Chao et al. \cite{chao2020real} applied a variant of the Maximum Entropy-based RL algorithm for model calibration of Turbofan Engines. In this work, the authors proposed the Lyapunov-Based Critic which to some extent provides stability guarantees which are essential for non-linear dynamical systems. We follow this approach here, by formulating Battery Model Calibration as the tracking problem.

\section{Method}

\subsection{Battery Discharge Model}

We follow Li-I Battery Model from NASA Prognostic Model Library \cite{daigle2013electrochemistry, nasaBattery}. It captures significant electro-chemical processes of the discharge and also models the effect of aging in terms of degradation parameters. However, the model needs to be provided with degradation parameters for accurate estimation of EOD time. The battery state is modeled by seven parameters, and changes over time as a function of input load and degradation state. Here, we just denote the state mathematically and refer the readers to the original paper \cite{daigle2013electrochemistry} for the physical meaning of these parameters. 
\begin{equation} \label{stateeq}
    \mathbf{x}(t) = [q_{s,p} \; \; q_{b,p} \; \; q_{b,n} \; \; q_{s,n} \; \; V_{o}^{'} \; \; V_{\eta,p}^{'} \; \; V_{\eta,n}^{'}]
\end{equation}

The input load at time t is $\mathbf{w}(t)$, and the model predicts the voltage $\mathbf{y}(t) = V$.

There are two main degradation parameters: (a) $q^{max}$ captures the decrease in active Lithium ions, and (b) $R_{o}$ captures the increase in internal resistance. These parameters are essential for the model dynamics, that are defined as follows:

\begin{equation}
    \begin{split}
        \mathbf{x}(t+1) & = f(\mathbf{x}(t), \mathbf{w}(t), q^{max}, R_{o}) \\
        \mathbf{y}(t+1) & = f(\mathbf{x}(t+1), q^{max}, R_{o}) 
    \end{split}
\end{equation}

Without any knowledge of the battery age, degradation parameters are initialized to ``perfect battery'' condition values, which are $q^{max} = 7600$, and $R_{o} = 0.117215$. Using these parameters, the model can estimate the initial state $\mathbf{x}(0)$. As the battery ages, $q^{max}$ decreases while $R_{o}$ increases. We try to infer these parameters by solving the state-tracking problem using RL. Here, we use the physics-based battery model as our Reinforcement Learning environment. However, in cases where such a model is difficult to obtain, it can be replaced by function approximators or surrogate models. 

\subsection{Markov Decision Process and Reinforcement Learning}
In this paper, we focus on the battery state tracking task which is modeled by a Markov decision process (MDP). An MDP can be described as a tuple, 
($S,A,c,P,\rho$), where $S$ is the set of states,
$A$ is the set of actions, $c (s,a)\in [0,\infty)$ is the cost function,and  $P (s'|s,a)$ is the transition probability function, and $\rho (s)$ is the starting state distribution. $\pi(a|s)$ is a policy denoting the probability of selecting action $a$ in state $s$. The state of a system at time $t$ is given by the state $s_t\in\mathcal{S}\subseteq \mathbb{R}^n$, where $\mathcal{S}$ denotes the state space. For our tracking strategy, we define the state at time t as $s_t \; = \; [\hat{\mathbf{x}}_{t},  \mathbf{x}_{t+1}, \mathbf{u}_{t+1}]$. Where, $\hat{\mathbf{x}}_t$ is the model predicted battery internal state and $\mathbf{x}_{t+1}$ is the real Battery state as described in eq.(\ref{stateeq}). The agent(calibrator) then controls the system's degradation parameters as an action $a_t \in\mathcal{A}\subseteq \mathbb{R}^m$ (e.g, $a_{t} = q^{max}$ or $R_{o}$) according to the policy $\pi(a_t|s_t)$, and resulting in the next state $s_{t+1}$. The transition of the state is computed by the battery model. The cost function $c(s_t,a_t) = ||\hat{\mathbf{x}}_{t+1} - \mathbf{x}_{t+1}||$ is a feedback signal to  the agent.

Our RL algorithm aims to find a 
policy $\pi$ which minimizes
$J_c(\pi) \doteq \mathbb{E}_{\tau \sim \pi}{\sum_{t=0}^{\infty} \gamma^t c(s_t, a_t)}$. Here, $\gamma \in [0,1)$ is the discount factor, $\tau$ denotes a trajectory ($\tau = (s_0, a_0, s_1, ...)$), and $\tau \sim \pi$ is shorthand for indicating that the distribution over trajectories depends on $\pi$: $s_0 \sim \rho$, $a_t \sim \pi(\cdot|s_t)$, $s_{t+1} \sim P(\cdot | s_t, a_t)$.

\subsection{Lyapunov-based actor-critic}
Since we target the state tracking task, we adopted the Lyapunov-based Actor-Critic (LAC). LAC has been proved to be able to learn policies with guaranteed stability, which is more capable and favorable of handling uncertainties compared to those without such guarantees in nonlinear control problems. 
Besides, LAC is based on the actor-critic maximum entropy framework \cite{haarnoja2018soft}, which can enhance the exploration of the policy and has been shown to substantially improve the robustness of the learned policy \cite{haarnoja2018soft}. LAC contains a Lyapunov critic function and a policy network. The Lyapunov critic plays an important role in both stability analysis and the learning of the actor. In view of the requirements in the LAC, we parameterized the Lyapunov candidate function as $L_c^{\phi}$ and construct the Lyapunov candidate by $L(s,a)=c+\max_{a'}\gamma L(s',a')$. During training, 
$L_c^{\phi}$ is updated to minimize the following objective function,
\begin{equation}
    J(L_c) = \mathbb{E}_{(s,a)\sim \mathcal{D}}\left[\frac{1}{2}(L_c(s,a)-L_c^{target}(s,a))^2\right]
\end{equation}
where $L_{target}$ is the approximation target related to the chosen Lyapunov candidate and D is the set of collected transition pairs. The approximation target is given by,
\begin{equation}
 L_{c}^{target} = c+\max_{a'}\gamma L_c^{\phi}(s',a')
\end{equation}

Then, based on the maximum entropy actor-critic framework, it uses the Lyapunov critic function in the policy gradient formulation. First, the objective of the policy network is summarized as follows:

\begin{equation}
\begin{aligned}
J(\pi) = &\mathbb{E}_{ \mathcal{D}}[ \beta [\log(\pi_\theta(f_\theta(\epsilon,s)|s))]+ \lambda(L_c((s',f_\theta(\epsilon,s')) - L_c(s,a)+\alpha_3c)]
\label{LAC}
\end{aligned}
\end{equation}
where $\pi_\theta$ is the policy parameterized by a neural network $f_\theta$, and $\epsilon$ is an input vector consisting of Gaussian noise. The $\mathcal{D}\doteq\{(s,a,s',c)\}$ is the replay buffer for storage of the MDP tuples. In the above objective, $\beta$ and $\gamma$ are positive Lagrange multipliers that control the relative importance of policy entropy versus the stability guarantee. And $\alpha_3$ is a constant for lyapunov energy decreasing objective. As in \cite{haarnoja2019soft}, the entropy of policy is expected to remain above the target entropy $\mathcal{H}_t$. The values of $\beta$ and $\lambda$ are adjusted through gradient method, thereby maximizing the objective:
\begin{gather}
J(\beta) = \beta\mathbb{E}_{(s,a)\sim\mathcal{D}}  [\log(\pi_\theta( a|s))+\mathcal{H}_t]\label{eq:temperature update}
\end{gather}
and the $\lambda$ is adjusted by the gradient method, thus maximizing the objective:
\begin{gather}
J(\lambda) = \lambda(L_c((s',f_\theta(\epsilon,s'))-L_c(s,a)+\alpha_3c)
\label{eq:gamma update}
\end{gather}

\subsection{Direct Mapping}
We also consider a simple fully connected neural network to learn a direct mapping from state $s_{t}$ to the degradation parameters ($a_{t}$ in RL setting).  However, this is a much simpler problem since it learns from the labelled pairs of "states" and "degradation parameters", which might not be easy to obtain in certain scenario (for every pairs of states and the underlying degradation, the assets need to be measured manually, which is considerable time consuming). Furthermore, the training dataset is required to be representative and cover all the different combinations to enable a reliable ML model. Also, in this setting state $s_{t}$ is formulated as  $s_t \; = \; [\mathbf{x}_{t},  \mathbf{x}_{t+1}, \mathbf{u}_{t+1}]$, and hence both $x_t$ and $x_{t+1}$ are real states. On the other hand, with RL parameter tracking formulation, we do not need such labelled training samples since an actor learns purely based on the rewards observed. Hence, we treat the results obtained with this supervised learning setup as an upper bound for our RL framework performance. 

\section{Experiments}

\subsection{Datasets and Model architecture}
We use the battery model from NASA Prognostic model library \cite{nasaBattery} to generate simulated data for our training process. As discussed earlier, we have two degradation parameters to calibrate in a battery model. We generate 5500 trajectories for each degradation parameter by varying the parameter value and load condition within a certain range. We consider each parameter degradation independently, hence, while varying one parameter, we keep the other one constant. Following the approach of \cite{daigle2016end}, we keep degradation parameters constant for a given discharge cycle. Also, we are assuming here that we the future battery load conditions are known for a discharge trajectory.

We use a fully connected neural network as a function approximator for our actor $f_\theta$ and Lyapunov critic, $L_c$. Both networks have 3 fully connected layers with 256 neurons each and LeakyReLU \cite{xu2015empirical} activation functions. For the policy network, we predict two values, mean and std, for each action. After this step, we use the squashed Gaussian policy \cite{haarnoja2018soft} to sample from the distribution. To ensure that the  Lyapunov values are positive, we use the sum-of-square of the final layer activations of the Lyapunov network as Lyapunov values. The parameters $\beta$ and $\lambda$ are also updated by using the loss defined in Eq.(\ref{eq:temperature update}) and Eq.(\ref{eq:gamma update}). We use Adam optimizer with the learning rate 5x10\textsuperscript{-4}.

For the Direct Mapping experiment, we use the same architecture as the policy network with the difference that only one output per action is learned since standard deviation is not required. We use the same optimizer and hyper-parameters as described above.

\subsection{Results}
For  model evaluation, we divide the dataset randomly in 70\% training and 30\% testing data 
We train RL model for one million steps while the direct mapping model has been trained for 30 epochs. 

\begin{figure}
\centering
\begin{subfigure}{.5\linewidth}
  \centering
  \includegraphics[width=\linewidth]{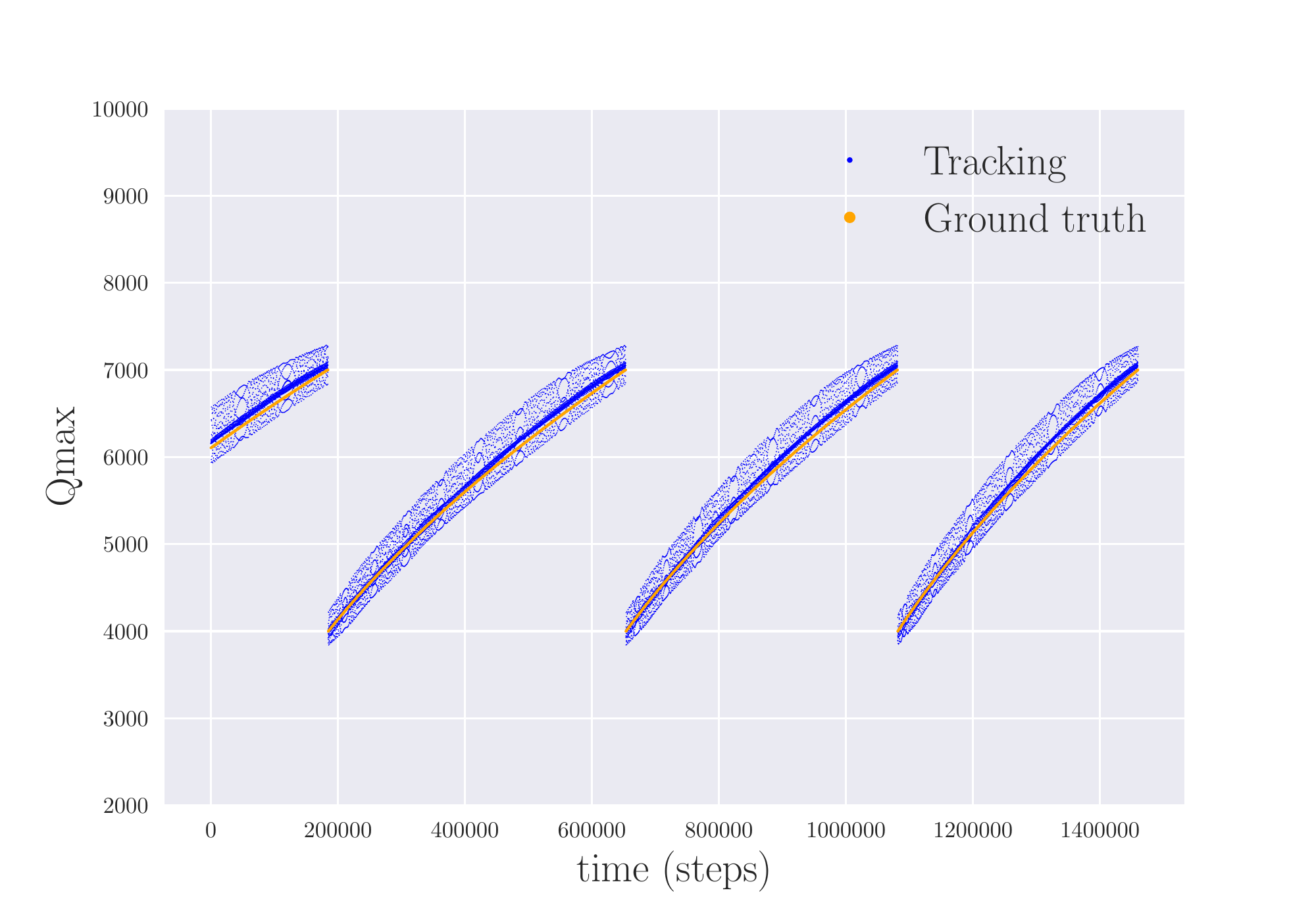}
  \caption{q\textsuperscript{max} tracking}
  \label{fig:RLq}
\end{subfigure}%
\begin{subfigure}{.5\linewidth}
  \centering
  \includegraphics[width=\linewidth]{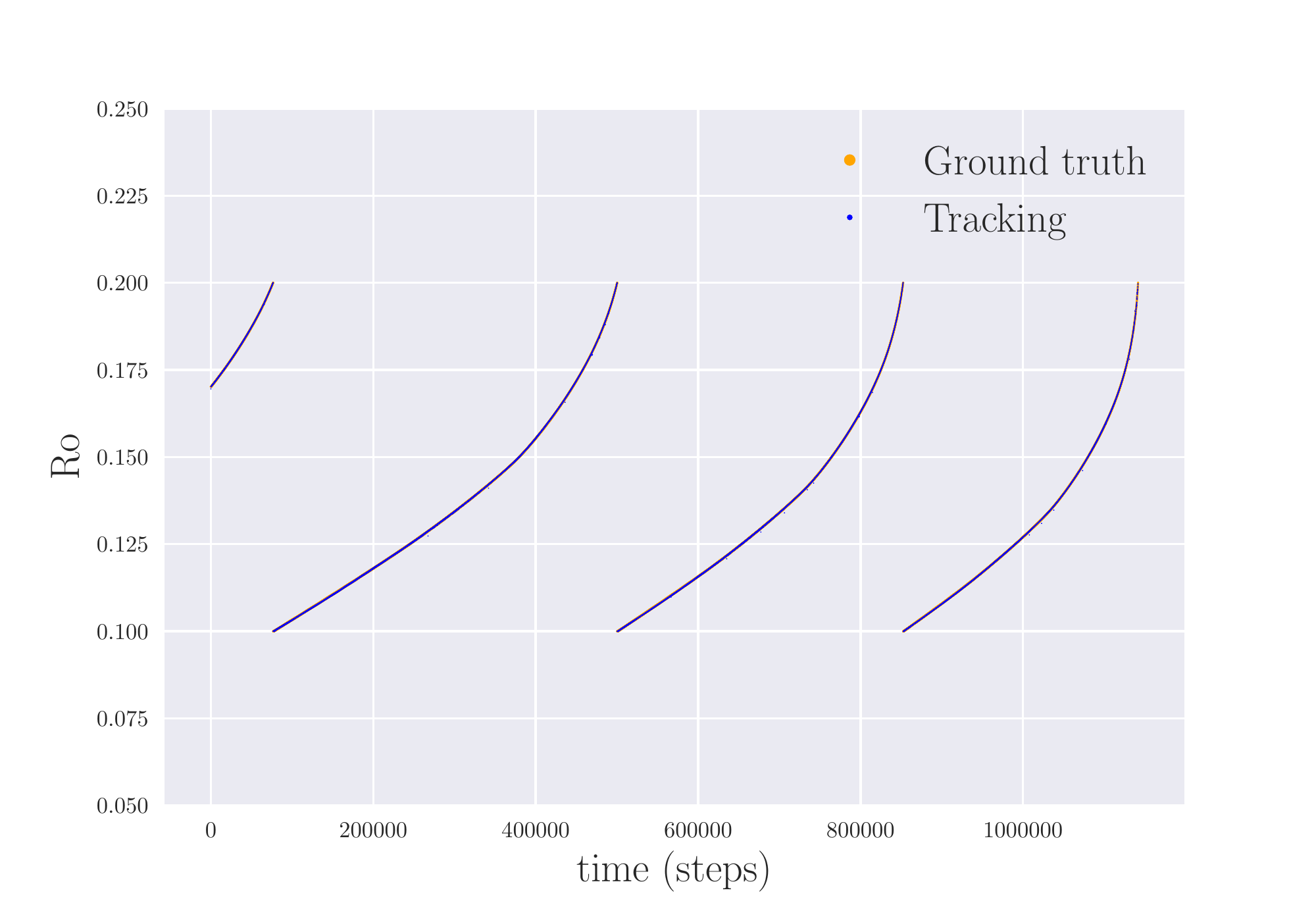}
  \caption{R\textsubscript{o} tracking}
  \label{fig:RLr}
\end{subfigure}
\caption{Battery model parameter calibration with RL}
\label{fig:RL}
\end{figure}

\begin{figure}
\centering
\begin{subfigure}{.5\linewidth}
  \centering
  \includegraphics[width=\linewidth]{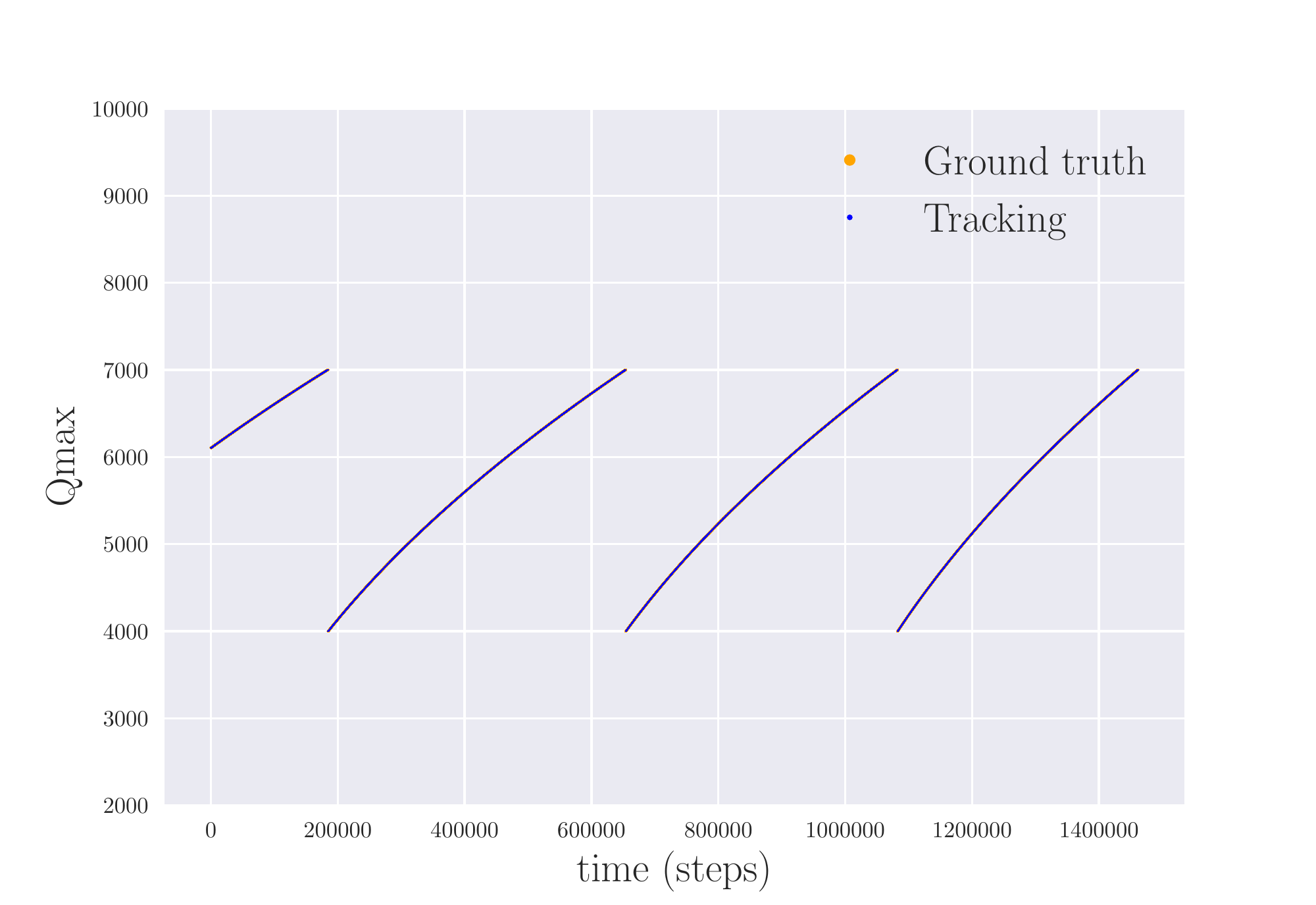}
  \caption{q\textsuperscript{max} tracking}
  \label{fig:DLq}
\end{subfigure}%
\begin{subfigure}{.5\linewidth}
  \centering
  \includegraphics[width=\linewidth]{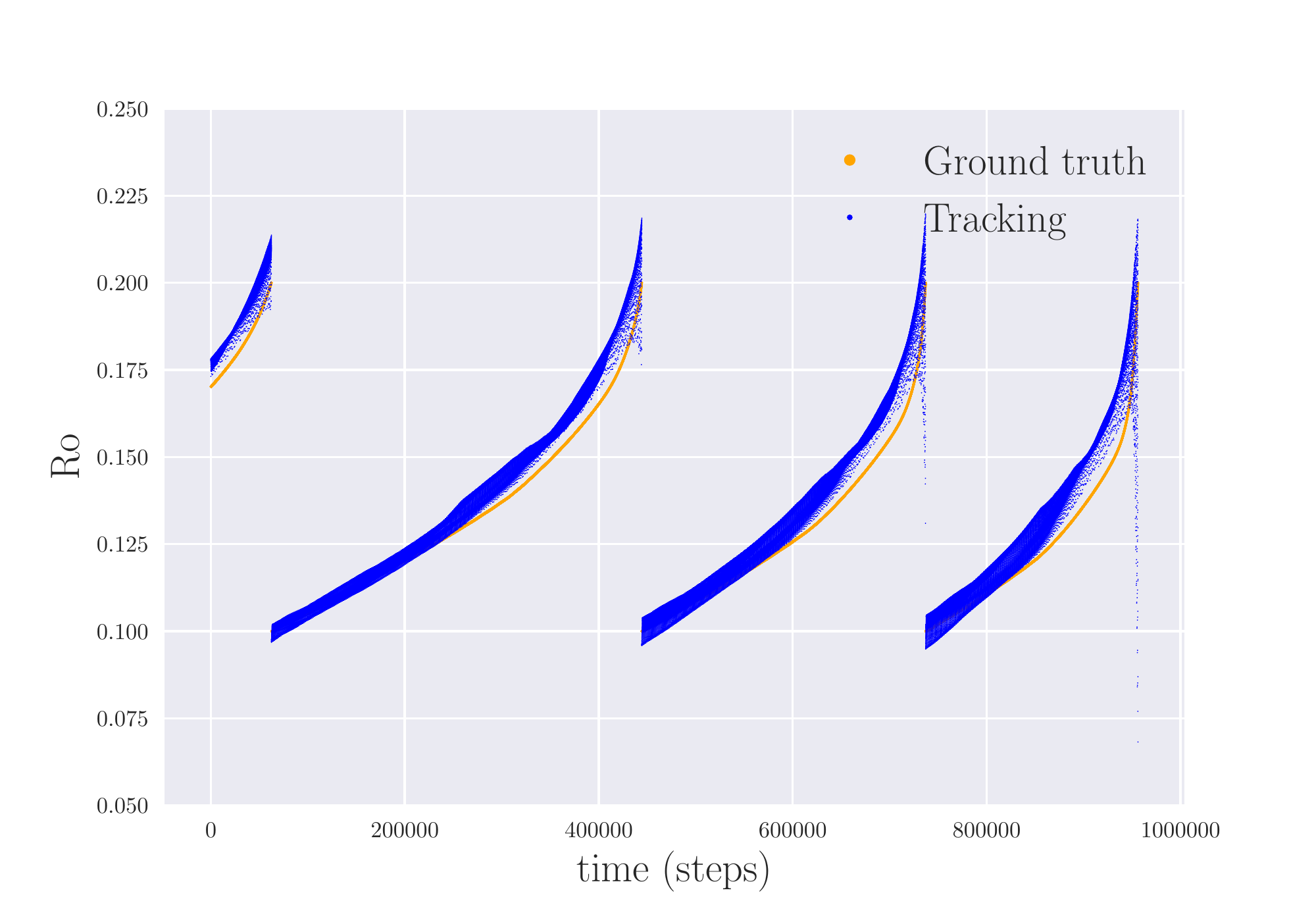}
  \caption{R\textsubscript{o} tracking}
  \label{fig:DLr}
\end{subfigure}
\caption{Direct Mapping of Battery state to degradation parameters}
\label{fig:DL}
\end{figure}

\subsubsection{RL Calibration}
We compare the inference accuracy of our RL-based approach with the direct mapping approach. As discussed earlier, direct mapping takes the labeled training pairs and hence, is expected to perform better than the indirect inference RL method. Inference trajectories of the degradation parameters are shown in Fig. (\ref{fig:RL}). Even though there is some variance in the inference of the parameter q\textsubscript{max}, we can see that most of the points are close to the true parameters, while in the case of R\textsubscript{o}, tracking works perfectly. We can observe bias in the direct mapping for R\textsubscript{o} parameter Fig.(\ref{fig:DLr}). The bias is much worse than the one in the indirect RL inference. Hence, the performance of the  RL method is either comparable or better as the direct mapping algorithm. This competitive performance is achieved while purely learning from the interactions and without any access to the ground truth.

\section{Conclusion}


In this paper, we present a new approach for the Battery model-calibration as a tracking problem. We solve this tracking problem by Lyapunov-based maximum entropy RL framework and show that the inference of this model provides accurate estimates of the model parameters. The performance of the RL framework is comparable to the supervised learning algorithm which requires labeled pairs of state observations and degradation parameters. The indirect inference as performed by the RL algorithm is a much much harder learning problem. Therefore, we propose a valid alternative for the scenarios where training data is limited. 

In the future, this method can be extended in the scenarios where the internal state of the model is not easy to obtain. For such cases, we can formulate the problem as a problem of tracking the output voltage. This is a much harder problem compared to the one analyzed here since RL has to learn the internal discharge model along with the degradation process purely from the observed rewards. Furthermore, in the current case study, we observed some variance in the inference of the parameter q\textsuperscript{max}. To tackle this, we can add a consistency penalty loss if two consecutive actions differ significantly. 

\nocite{*}
\bibliographystyle{IEEEtranN}
\bibliography{ml4eng}
\addcontentsline{toc}{chapter}{Bibliography}
\cleardoublepage

\end{document}